\documentclass[
11pt,
a4paper,
oneside, 
headinclude,footinclude, 
BCOR5mm,
]{article}

\usepackage{tgtermes}

\usepackage[utf8]{inputenc}
\usepackage{amsmath}
\usepackage{subcaption}
\usepackage{caption}
\usepackage{longtable}
\usepackage{amsthm}
\usepackage{color}
\usepackage{sectsty}
\usepackage{hyperref}
\hypersetup{
    colorlinks=true, % make the links colored
    linkcolor=blue, % color TOC links in blue
    urlcolor=red, % color URLs in red
    linktoc=all % 'all' will create links for everything in the TOC
}
\usepackage{amssymb}
\usepackage{pagecolor,lipsum}
\usepackage{enumerate}
\usepackage{pdfpages}
\usepackage{xcolor}
\usepackage{tikzsymbols}
\usepackage{blkarray}
\usepackage{enumitem}
\usepackage{graphicx}
\usepackage{setspace}
\usepackage{authblk}
\usepackage{soul}
\usepackage{enumitem}
\usepackage{soul}
\usepackage{MnSymbol}
\usepackage{mathtools}
\usepackage{braket}
\usepackage{mdframed}
\usepackage{amsmath}
\usepackage[thinc]{esdiff}
\usepackage{array}
\usepackage{booktabs}
\usepackage[linesnumbered,ruled,vlined,english]{algorithm2e}
\usepackage[noend]{algpseudocode}
\usepackage[linesnumbered,ruled,vlined,english]{algorithm2e}
\usepackage{geometry}
 \let\oldpara\paragraph
\renewcommand{\paragraph}[1]{\vspace{-0.5cm}\oldpara{#1}}

\newcommand{\cnp}{\mbox{\textbf{NP}}}

\newcommand{\mc}[1]{\mathcal{#1}}

\newcommand{\ep}[1]{\mathbb{E}}
\newcommand{\done}[1]{\textcolor{blue}{(\textbf{Done})}}

\newcommand{\pr}[1]{\mathbb{Pr}}

\newcommand{\innerthmname}{}% initialize

\theoremstyle{definition}

\newcommand{\z}[1]{\mathbb{Z}}
\newcommand{\nminfpe}[1]{\textsc{NMin-FPE}}
\newcommand{\nmaxfpe}[1]{\textsc{NMax-FPE}}
\newcommand{\nminfpr}[1]{\textsc{NMin-FPR}}
\newcommand{\nmaxfpr}[1]{\textsc{NMax-FPR}}

\newcommand{\zt}[1]{\mathbb{Z}_T}

\newcommand{\G}{\mathcal{G}}
\newcommand{\Gv}{\mathcal{V}}
\newcommand{\Ge}{\mathcal{E}}

\newcommand{\Ndim}[1]{\text{Ndim}(#1)}

\newcommand{\sampcom}[1]{m_{#1}(\delta, \epsilon)}

\newcommand{\pac}{\textsf{PAC}}
\newcommand{\hclass}{\mathcal{H}}

\usepackage{float}

\newcommand{\crp}{\textbf{RP}}

\newcommand{\yes}{\texttt{YES}}

\newcommand{\QED}{\hfill\rule{2mm}{2mm}}

\newcommand{\cals}{\mbox{$\mathcal{S}$}}

\newtheorem{theorem}{\textbf{Theorem}}[section]
\newtheorem{corollary}{\textbf{Corollary}}[theorem]

\newtheorem{lemma}[theorem]{\textbf{Lemma}}

\newtheorem{observation}[theorem]{Observation}

\newcommand{\calc}{\mbox{$\mathcal{C}$}}
\newcommand{\calcp}{\mbox{$\mathcal{C'}$}}

\newcommand{\bbo}{\mbox{$\mathbb{O}$}}

\newcommand{\obset}{\mbox{$\mathbb{O}$}}

\newcommand{\bbb}{\mbox{$\mathbb{B}$}}

\newcommand{\undthr}{(\textsc{Undir},~\textsc{Thresh})-SyDSs}

\newcommand{\dthr}{(\textsc{Dir}, $\Delta$, \textsc{Thresh})-SyDSs}

     \newcommand{\matthr}{(\textsc{Match},\textsc{Thresh})-SyDSs} 

\newcommand{\treethrtwo}
      {(\textsc{Tree},~\textsc{Thresh}$\,=2$)-SyDSs}

\newcommand{\hclasspartial}{\mathcal{H}_{\textsc{obs}}}

\usepackage{bigints}
\DeclareMathOperator{\sac}{\mathcal{M}}
\newcommand{\davg}{d_\text{avg}}

\newcommand{\undsym}{(\textsc{Undir},~\textsc{Symm})-SyDSs}

\newcommand{\M}{\mathcal{M}}

\newcommand{\hclassmatch}{\mathcal{H}_{\textsc{match}}}
\newcommand{\hclassdb}{\mathcal{H}_{\textsc{dir,$\Delta$}-\textsc{bounded}}}

\newcommand{\Gobs}{\mbox{$\mathcal{G}_\text{obs}$}}
\newcommand{\Gm}{\mbox{$\mathcal{G}_m$}}
\newcommand{\Aerm}{\mbox{$\mathcal{A}_{\textsf{ERM}}$}}
\newcommand{\Apac}{\mbox{$\mathcal{A}_{\textsf{PAC}}$}}

\newcommand{\Ierm}{\mbox{$\mathcal{I}_{\textsf{ERM}}$}}
\newcommand{\Ipac}{\mbox{$\mathcal{I}_{\textsf{PAC}}$}}

\newcommand{\gclass}{\mbox{$\mathcal{G}$}}

\definecolor{mycolor}{rgb}{0.122, 0.435, 0.698}
\newmdenv[topline=false, bottomline=false, rightline=false, innerlinewidth=0.4pt, roundcorner=4pt,linecolor=black,innerleftmargin=6pt,
innerrightmargin=6pt,innertopmargin=1pt,innerbottommargin=6pt]{mybox}

\newmdenv[backgroundcolor=gray!10, topline=false, bottomline=false, rightline=false, innerlinewidth=0.4pt, roundcorner=4pt,linecolor=black,innerleftmargin=6pt,
innerrightmargin=6pt,innertopmargin=3pt,innerbottommargin=6pt]{mybox2}

\newmdenv[backgroundcolor=blue!5, topline=false, bottomline=false, rightline=false, leftline=false, innerlinewidth=0.4pt, roundcorner=4pt,innerleftmargin=10pt,
innerrightmargin=10pt,innertopmargin=10pt,innerbottommargin=10pt]{mybox3}

\definecolor{darkblue}{RGB}{0,0,76}
\title{\textbf{Learning the Topology and Behavior of Discrete \\ Dynamical Systems}}

\author { \small
    Zirou Qiu,\textsuperscript{1,2}
    Abhijin Adiga, \textsuperscript{2}
    Madhav V. Marathe,\textsuperscript{1,2}
    S. S. Ravi,\textsuperscript{2,3}\\
    Daniel J. Rosenkrantz,\textsuperscript{2,3}
    Richard E. Stearns,\textsuperscript{2,3}
    Anil Vullikanti\textsuperscript{1,2}
}

\affil[1]{\small Computer Science Dept., University of Virginia.}
\affil[2]{\small Biocomplexity Institute and Initiative, University of Virginia.}
\affil[3]{\small Computer Science Dept., University at Albany – SUNY.}

\date{}

\begin{document}
\maketitle

% --------------------
%       Content      -
% --------------------
\vspace{-1cm}
\begin{abstract}

\noindent
Discrete dynamical systems are commonly used to model the spread of contagions on real-world networks. Under the \pac{} framework, existing research has studied the problem of learning the behavior of a system, assuming that the underlying network is known. In this work, we focus on a more challenging setting: to learn {\em both the behavior and the underlying topology} of a black-box system. We show that, in general, this learning problem is computationally intractable. On the positive side, we present efficient learning methods under the \pac{} model when the underlying graph of the dynamical system belongs to some classes. Further, we examine a relaxed setting where the topology of an unknown system is partially observed. For this case, we develop an efficient \pac{} learner to infer the system and establish the sample complexity.  Lastly, we present a formal analysis of the expressive power of the hypothesis class of dynamical systems where both the topology and behavior are unknown, using the well-known formalism of the Natarajan dimension. Our results provide a theoretical foundation for learning both the behavior and topology of discrete dynamical systems.

\smallskip
\noindent
\textbf{Conference version.} The conference version of the paper is accepted at \texttt{\textbf{AAAI-2024}}: \href{https://ojs.aaai.org/index.php/AAAI/article/view/29390}{\textbf{Link}}. 
\end{abstract}

\section{Introduction}
\label{sec:intro}
Discrete dynamical systems are formal models for numerous cascade processes, such as the spread of social behaviors, information, diseases, and biological phenomena~\cite{battiston2020networks, ji2017mathematical, cohen2010quorum,bailon:nsr-2011,romero:www-2011,sabhapandit2002hysteresis}. A discrete dynamical system consists of an \textit{underlying graph} with vertices representing entities (e.g., individuals, genes), and edges representing connections between the entities. Further, each vertex has a contagion \textit{state} and an \textit{interaction function} (i.e., behavior), which specify how the state changes over time. Overall, vertices update states using interaction functions as the system dynamics proceeds in discrete time.

Due to the large scale of real-world cascades, a complete specification of the underlying dynamical system is often not available. To this end, learning the unknown components of a system is an active area of research~\cite{adiga2019pac,chen2022learning,
conitzer2020learning,dawkins2021diffusion}. One ongoing line of work is to infer the unknown {\em interaction functions} or the {\em topology} of a system. Interaction functions and the network topology play critical roles in the system dynamics. 
The topology encodes the underlying relationships between the entities, while the interaction functions provide the decision rules that entities employ to update their contagion states. 
The class of threshold interaction functions~\cite{granovetter-1978}, which are widely used to model the spread of social contagions~\cite{watts2002simple}, is one such
example. 
Specifically, each entity in the network has a decision threshold that represents the tipping point for a behavioral (i.e., state) change. In the case of a rumor, a person's belief shifts when the number of neighbors believing in the rumor reaches a certain threshold.

Previous research~\cite{Adiga-etal-2017} has presented efficient algorithms for inferring classes of interactions functions (e.g., threshold functions)   based on the observed dynamics, {\em assuming that the underlying network is known}. To our knowledge, the more challenging problem of learning a system from
observed dynamics, where \textbf{both} the interaction function and the topology are \textit{unknown}, has not been examined. In this work, we fill this gap with a theoretical study of the problem of \textbf{learning both the network and the interaction functions of an unknown dynamical system}. 

\noindent
\textbf{Problem description.} Consider a black-box networked system where both the interaction functions and the network topology are \textit{unknown}. Our objective is to recover a system that captures the behavior of the true but unknown  system while providing performance guarantees under the \textsf{P}robably \textsf{A}pproximately \textsf{C}orrect (\pac{}) model~\cite{valiant:acm1984}. 
We learn from \textit{snapshots of the true system's dynamics}, a common scheme considered in related papers~(e.g., \cite{chen2021network, conitzer2020learning, wilinski2021prediction}). 
Since our problem setting also involves multiclass learning, we examine the {\em Natarajan dimension}~\cite{natarajan1989learning}, a well-known generalization of the VC dimension~\cite{vapnik2015uniform}. In particular, the Natarajan dimension quantifies the expressive power of the hypothesis class and characterizes the sample complexity of learning. Overall, we aim to answer the following two questions: $(i)$ {\em Can one efficiently learn the black-box system, and if so, how many training examples are sufficient?} $(ii)$ {\em What is the expressive power of the hypothesis class of networked systems?}

The key challenges of our learning problem arise from the 
incomplete knowledge of the network topology and the interaction functions of the nodes in the system.
For example, when we consider systems whose underlying graphs are undirected and the interaction functions are Boolean threshold
functions, the number of potential systems is
$\Theta(2^{\binom{2}{2}} \cdot  n^n)$, where $n$ is the
number of nodes in the underlying graph.
Therefore, a learner needs to find an appropriate hypothesis in a very large space. Further, in general, the training set (which consists of snapshots of the system dynamics) may not contain sufficient information to recover the underlying network structure efficiently (as we show that the problem is computationally intractable). 
\textbf{Our main contributions are summarized below}.

%\begin{itemize}[leftmargin=*,noitemsep,topsep=0pt]
\smallskip

\noindent 
1. \textbf{Hardness of \pac{} learning}:~ We show that
in general, hypothesis classes corresponding to dynamical 
system, where both the network topology and the interaction functions are unknown, are not efficiently \pac{} learnable unless\footnote{For information regarding the complexity
classes \cnp{} and \crp{}, we refer the reader
to \cite{Arora-Barak-2009}.} $\textbf{NP} = \textbf{RP}$.
We prove this result by first formulating a 
suitable decision problem and showing that the problem
is \cnp-complete. We use the hardness result for the 
decision problem to establish the hardness 
of \pac{} learning. Our results show that the
learning problem remains hard for several classes
of dynamical systems (e.g., systems on undirected
graphs with threshold interaction functions).

\smallskip

\noindent
2. \textbf{Efficient \pac{} learning algorithms for
special classes.}~ In contrast with the general hardness result above, we identify some special classes of systems
which are efficiently \pac{} learnable.
The two classes that we identify correspond to systems
on directed graphs with bounded indegree and those where
underlying graph is a perfect matching. In both cases,
the interaction functions are from the class of threshold functions.
These results are obtained by showing that these systems have efficient consistent learners 
and then appealing to the known
result \cite{Shwartz-David-2014} that hypothesis classes for which there are efficient
consistent learners are also efficiently \pac{} learnable.

\smallskip 

 \noindent 
3. \textbf{Learning under a relaxed setting.}~
We consider a relaxed setting where the underlying network is partially observed. Towards this end, we present an efficient \pac{} learner and establish the sample complexity of learning under this setting.

\smallskip

\noindent 
4. \textbf{Expressive power.}~ We present an analysis of the Natarajan dimension (Ndim), which measures the expressive power of the hypothesis class of networked systems. In particular, we give a construction of a shatterable set and prove that the Ndim of the hypothesis class is at least $\lfloor n^2/4 \rfloor$, where $n$ is the number of vertices. Further, we show that the upper bound on the Ndim is $O(n^2)$. Thus, our lower bound is tight to within a constant factor. Our result also provides a lower bound on the sample complexity. 

\smallskip

\noindent 
\textbf{Related work.} Learning unknown components of networked systems is an active area of research. Many researchers have studied the problem of identifying missing components (e.g., learning the diffusion functions at vertices, edge parameters, source vertices, and contagion states of vertices) in contagion models by observing the propagation dynamics. For instance, Chen et al.~\cite{chen2021network} infer the edge probability and source vertices under the independent cascade model. 
Conitzer et al.~\cite{conitzer2020learning} investigate the problem of inferring opinions (states) of vertices in stochastic cascades under the \pac{} scheme. Lokhov \cite{lokhov2016reconstructing} studies the problem of reconstructing the parameters of a diffusion model given infection cascades. Inferring threshold functions of vertices from social media data is also considered in \cite{bailon:nsr-2011,romero:www-2011}. Learning the source vertices of infection
for contagion spreading is addressed in
\cite{zhu:aaai2017,shah2011rumors}. 
The problem of inferring the network structure has also been studied, see, for example \cite{huang2019statistical, pouget2015inferring, abrahao2013trace,gomez2010inferring, 
soundarajan2010recovering}.
The problem of inferring the network structure and
the interaction functions of dynamical systems has been
studied under a different model in~\cite{RAM+2022},
where a user can submit queries to the system and obtain
responses. This is very different from our model where a
learning algorithm must rely on a given set of observations and cannot interact with the unknown system.

The work that is closest to ours is 
\cite{Adiga-etal-2017}, where
the problem of inferring the interaction
functions in a system from observations is considered, {\em under the assumption
that the network is known}. For the type of observations used in our paper, it is shown in \cite{Adiga-etal-2017} that the local function inference problem can be solved efficiently.
However, the techniques used in~\cite{Adiga-etal-2017} are not applicable to our context, since 
the network is unknown.

\section{Preliminaries}
\label{sec:prelim}
\subsection{Discrete Dynamical Systems} \label{sse:syds_def}
A {\em Synchronous Dynamical System} (\textbf{SyDS}) \cals{} over the Boolean domain $\bbb{} = \{0,1\}$ is a pair ${\cal S}  = (\mc{G}, {\cal F})$, where:

\begin{itemize}[leftmargin=*,noitemsep,topsep=0pt]
\item $\G{} = (\Gv{},\Ge{})$ is the underlying undirected graph. We let $n = |\Gv{}|$. Each vertex in $\Gv{}$ has a {\em state} from \bbb, representing its contagion state (e.g., inactive or active). 
\item ${\cal F} = \{f_1, f_2, \ldots, f_n \}$ is a collection of functions, where $f_i$ is the {\em interaction function} of vertex $v_i \in \Gv{}$, $1 \leq i \leq n$.
\end{itemize}

\smallskip
\noindent
\textbf{Interaction functions.} \label{sse:special_bool} The system evolves in {\em discrete} time steps, with vertices updating their states \emph{synchronously} using the interaction functions. 
For a graph $\G{}$ and a vertex $v$, we let $N(\G{}, v)$ and $N^+(\G{}, v)$ denote the open and closed neighborhoods\footnote{The open neighborhood of a node $v$ contains each node $u$ such that $\{u,v\}$ is an
edge in $\G${}. The closed neighborhood of $v$ includes
$v$ and all the nodes in its closed neighborhood.}
of $v$, respectively.
At each time step, the inputs to the interaction function $f_i \in \mathcal{F}$ are the states of the vertices in $N^+(\G{}, v_i)$; $f_i$ computes the next state of $v_i$.

\begin{figure}[!h]
%\rule{\textwidth}{0.01in}
\begin{center}
\includegraphics[width=0.7\textwidth]{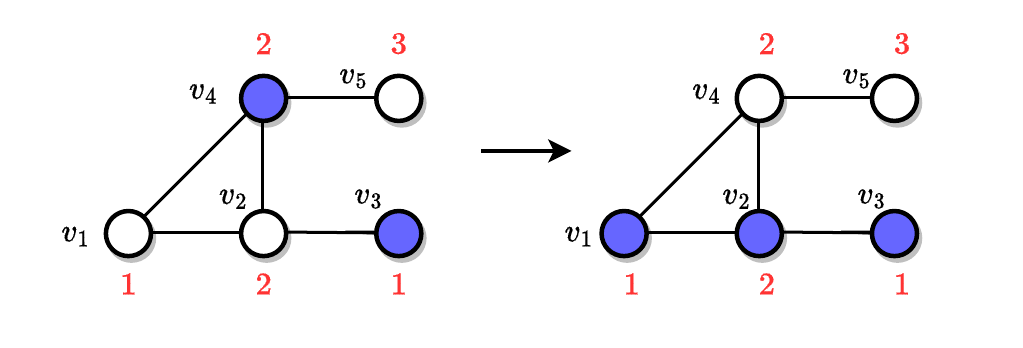}
\end{center}
\caption{An example of a single transition of a SyDS with threshold interaction functions. The threshold values of vertices are highlighted in red. State-1 vertices are in blue.}
\label{fig:syds_example}
\end{figure}

Our work focuses on the class of \textbf{threshold} interaction functions, which is a classic framework modeling the spread of social contagions~\cite{watts1998collective, granovetter-1978}. Specifically, each vertex $v_i \in \Gv{}$ has an integer threshold $\tau_{v_i} \geq 0$, and $f_i$ outputs $1$ if the number of 1's in the input to  $f_i$ (i.e., the number of state-1 vertices in $v_i$'s closed neighborhood) is at least $\tau_{v_i}$; $f_i$ outputs $0$ otherwise.
An example of a SyDS is shown in Figure~\ref{fig:syds_example}.

A \textbf{configuration} \calc{} of a SyDS $\cals{}$ at a given time step is a binary vector of length $n$ that specifies the states of all the vertices at that time step. Let $\mc{X} = \{0, 1\}^n$ be the set of all configurations. 
For a given vertex $v$, let $\calc(v)$ denote the state of $v$ in $\calc$, and for a given vertex set $\mc{Y} \subseteq \Gv{}$, let $\calc[\mc{Y}]$ denote the projection of \calc{} onto $\mc{Y}$.
If the system $\mc{S}$ evolves from \calc{} to a configuration $\calc{}'$ in one step, denoted by $\mc{S}(\mc{C}) = \calc{}'$, then $\calc{}'$ is called the \textbf{successor} of \calc{}. 
Since the interaction functions considered in our work  are deterministic, each configuration has a unique successor.
For a given configuration $\calc$ and a vertex set $\mc{Y}$, we let $score(\calc, \mc{Y})$ denote the number of vertices $v \in \mc{Y}$ such that $\calc(v) = 1$. 

\noindent
\textbf{SyDSs over directed graphs.}
SyDSs can also be defined over \emph{directed} graphs. In such a case, the inputs to the local function at a vertex $v_i$ are the state of $v_i$ and those of the in-neighbors of $v_i$ (i.e., vertices from which $v_i$ has incoming edges).
All the other definitions for SyDSs on directed graphs are the same as those for SyDSs on undirected graphs. In this paper, we assume that the underlying graph is {\em undirected}, unless specified otherwise.

%% Added by Ravi.
\newcommand{\gtruth}{\mbox{${\cal S}^*$}}

\subsection{The Learning Problem}
\label{sse:addl_def}

\textbf{Hypotheses.} Let ${\cal S}^* = (\G{}, {\cal F})$ be a ground truth SyDS.
The learner does \emph{not} know either the graph $\G{}$ or the set of functions ${\cal F}$.
Other than the training set, the only information provided to the learner consists of the number $n$ of vertices, the class of interaction functions, and the class of the underlying graph.
The learner must find a \emph{hypothesis} consisting of an underlying graph (from the given graph class) and an interaction function (from the given class of interaction functions) 
for each node of the graph from the set of all such
hypotheses.
As an  illustration,
let $\hclass{}$
% H({Undir,Thr})
denote the {\em hypothesis class} where the underlying graph of \gtruth{} is undirected, and the interaction functions are threshold functions.
Since there are $2^{\binom{n}{2}}$ undirected graphs with $n$ nodes and each node may have
$\Theta(n)$ choices for threshold values, the
 {\em hypothesis class} $\hclass{}$ 
has $\Theta(2^{\binom{n}{2}} n^n)$ 
systems (i.e., hypotheses).
%over all the possible graphs structures and %threshold values. 
The learner aims to infer a system $\mc{S} \in \hclass{}$ that is close to the true system $\mc{S}^*$ by recovering the graph and the interaction functions of $\mc{S}$.

\smallskip
\noindent
\textbf{Learning setting.} Our algorithms learn the ground truth system ${\cal S}^*$ from its
observed dynamics.
Let $\bbo{} = \{(\mc{C}_i, \mc{C}'_i)\}_{i = 1}^{q}$ be a {\em training set} of $q$ examples, which consists of the snapshots of the dynamics of \gtruth{} in the form of configuration-successor pairs. Specifically, $\mc{C}_i$ is a drawn i.i.d. from a distribution ({\em unknown} to the learner) $\mc{D}$ over $\mc{X}$, and $\mc{C}_i' = \mc{S}^*(\mc{C}_i)$ is the successor of $\mc{C}_i$ under $\mc{S}^*$. Let $\bbo \sim \mc{D}^q$ denote such a training set. We say that a hypothesis $\mc{S}$ is \textbf{consistent} with $\bbo{}$ if 
$\forall (\mc{C}_i, \mc{C}_i') \in \bbo{}$,~ $\mc{S}(\mc{C}_i) = \mc{C}_i'$. 
For each vertex $v \in \Gv{}$, we define a partition of $\bbo{}$ into two subsets $\{\bbo{}_{0,v}, \bbo{}_{1,v}\}$, such that $\mc{C}'(v) = i$, for all $(\mc{C}, \mc{C}') \in \bbo{}_{i,v}$, $i = 0, 1$. Note that in general, the hypothesis being learned is a successor function. 

%In our context, such a hypothesis is chosen from $\hclass{}$ in the %form of a dynamical system, where both the graph and the interaction %functions are specified.
For simplicity,
we present the necessary definitions for the \pac{}  model
(see e.g.,~\cite{Shwartz-David-2014})
using $\hclass{}$. 
These definitions also apply to other hypothesis
classes considered in this paper.
The \textbf{true error} of a hypothesis $\mc{S} \in \hclass{}$ is defined as $L_{(\mc{D}, \mc{S}^*)}(\mc{S}) := \Pr{}_{\mc{C} \sim \mc{D}} [\mc{S}(\mc{C}) \neq \mc{S}^*(\mc{C})]$. In the \pac{} model, the goal is to infer a hypothesis $\mc{S} \in \hclass{}$ s.t. with probability at least $1 - \delta$ over $\bbo{} \sim \mc{D}^q$, the true error $L_{(\mc{D}, \mc{S}^*)}(\bbo{}) \leq \epsilon$, for any given $\epsilon, \delta \in (0, 1)$. The class $\hclass$ is {\em efficiently \pac{} learnable} if such an  $\mc{S} \in \hclass{}$ can be inferred in polynomial time (w.r.t. $n$, $1/\delta$ and $1/\epsilon$). The minimum number of training examples needed to achieve the above guarantee is called the \textbf{sample complexity} of learning $\hclass{}$.

\smallskip
\noindent
% \textbf{Natarajan dimension.} A hypothesis $\mc{S}$ maps a configuration $\mc{C}$ to one 
% of the $2^n$ configurations (classes). Thus, our problem involves multiclass classification. The Natarajan dimension~\cite{natarajan1989learning} is a generalization of the VC dimension to the multiclass setting, which measures the {\em expressive power} of the hypothesis class $\hclass{}$. Formally, a set $\mc{R}$ of configurations is \textbf{shattered} by $\hclass{}$ if there exists two functions $g_1, g_2 : \mc{X} \rightarrow \mc{X}$, such that for every $\mc{C} \in \mc{R}$, $(i)$ $g_1(\mc{C}) \neq g_2(\mc{C})$, and $(ii)$ for every subset $\mc{R}' \subseteq \mc{R}$, there exists $\mc{S} \in \hclass{}$ such that $\forall \mc{C} \in \mc{R}', \; \mc{S}(\mc{C}) = g_1(\mc{C})$ and $\forall \mc{C} \in \mc{R} \setminus \mc{R}', \; \mc{S}(\mc{C}) = g_2(\mc{C})$. The \textbf{Natarajan dimension} of $\hclass{}$, denoted by $\Ndim{\hclass{}}$, is the \emph{maximum} size of a shatterable set. 
% Overall, a larger $\Ndim{\hclass{}}$ implies a higher the expressive power of $\hclass{}$.
\textbf{Natarajan dimension.} 
A hypothesis $\mc{S}$ can be viewed as a function $\mc{X} \rightarrow \mc{X}$, where each of the $2^n$ possible target configurations is considered a class.
Thus, our learning problem involves multiclass classification. 
The Natarajan dimension~\cite{natarajan1989learning} is a generalization of the concept of VC dimension to a multiclass setting, 
and measures the {\em expressive power} of a given hypothesis class $\hclass{}$. 
Formally, a set $\mc{R}$ of configurations is \textbf{shattered} by $\hclass{}$ if there exists two functions $g_1, g_2 : \mc{X} \rightarrow \mc{X}$, such that:
$(i)$ for every $\mc{C} \in \mc{R}$, $g_1(\mc{C}) \neq g_2(\mc{C})$, 
and $(ii)$ for every subset $\mc{R}' \subseteq \mc{R}$, there exists $\mc{S} \in \hclass{}$ such that $\forall \mc{C} \in \mc{R}', \; \mc{S}(\mc{C}) = g_1(\mc{C})$ and $\forall \mc{C} \in \mc{R} \setminus \mc{R}', \; \mc{S}(\mc{C}) = g_2(\mc{C})$. The \textbf{Natarajan dimension} of $\hclass{}$, denoted by $\Ndim{\hclass{}}$, is the \emph{maximum} size of a shatterable set. 
%Overall, a larger $\Ndim{\hclass{}}$ implies a higher expressive power %of $\hclass{}$.
In general, the larger the Natarajan dimension of a hypothesis class, the higher its expressive power.

\section{\pac{} Learnability and Sample Complexity}\label{sec:general_complexity}

We first establish the {\em intractability} of 
efficiently \pac{} learning threshold dynamical systems over general graphs. We then present {\em efficient} algorithms for learning systems over several graph classes. We also analyze the {\em sample complexity} of learning the corresponding hypothesis classes.

\subsection{Intractability of \pac{} learning}
To establish the hardness of learning, we first formulate a decision problem for SyDSs and show that the problem is \cnp-hard. We use this to establish a general hardness result 
under the \pac{} model. We define restricted classes of SyDSs by specifying constraints on the underlying graph and the interaction functions. For example, we use 
the notation \undthr{} to denote the restricted class of SyDSs whose $(i)$ underlying graphs are undirected and $(ii)$ interaction functions are threshold functions. Several other restricted classes of SyDSs will be considered in this section.

Given a set \obset{} of observations, with each observation being a pair of the form $(\calc{}, \calcp{})$ where $\calcp${} is the successor of $\calc${}, we say that a SyDS \cals{} is \textbf{consistent} with \obset{} if for each observation $(\calc, \calcp) \in \obset$, the successor of $\calc$ produced by \cals{} is \calcp{}. A basic  decision problem in this context is the following: given a set \bbo{} of observations,  determine whether there is a SyDS that is  consistent with \bbo{}. One may also want the consistent SyDS to be in a restricted class $\Gamma$. We refer to this problem as the $\Gamma$-\textbf{Consistency} problem:

 \smallskip 
\begin{mybox2}
\textbf{$\Gamma$-Consistency}

\noindent
\underline{Instance}: A vertex set $\Gv{}$ and a set of observations
\obset{} over $\Gv{}$.

\noindent
\underline{Question}: Is there a SyDS in the class $\Gamma$ that is consistent with \obset{}?
\end{mybox2}

\noindent
\textbf{Hardness.} The following lemma shows the hardness
of $\Gamma$-consistency for two restricted classes of SyDSs: 

\noindent
$(i)$ \undthr{}, where the graph is undirected
and interaction functions are threshold functions.

\noindent
$(ii)$ \treethrtwo{}, where the graph is a tree and each interaction function is a 2-threshold function.  
%For space reasons, provide a proof sketch for $(i)$ here.
A full proof of the lemma appears in the appendix.

\begin{mybox2}
\begin{lemma}\label{lem:np-hardness}
The $\Gamma$-Consistency problem is \textbf{NP}-complete for
the following classes of SyDSs: \\ $(i)$ \undthr{} and $(ii)$ \treethrtwo{}.
\iffalse 
%%%%%%%%%%%%%%%%
and
(c) \undsym{} ((where the underlying graph is undirected
and each interaction function is 
a symmetric\footnote{A symmetric Boolean function 
depends only on the number of 1's in the input.
Examples of symmetric functions are AND, OR and threshold
functions.}
function).
%%%%%%%%%%%%
\fi
\end{lemma}
\end{mybox2}

\iffalse 
%%%%%%%%%%%%
\begin{mybox2}
\begin{lemma}
The problem $\Gamma$-Consistent is $\cnp$-complete for the class
\treethrtwo{}, even under the restriction that there is exactly one observed transition that is not a fixed point transition.
\end{lemma}
\end{mybox2}

\noindent
\textit{Remark.} We can further show that the problem is also hard for the more general class of {\em symmetric} functions. See the Appendix for details.
%%%%%%%%%
\fi

\noindent 
\textit{Proof sketch for $(i)$:}~
It can be seen that the problem is in \cnp{}.
The proof of $\cnp$-hardness is via a reduction from 3SAT.
Let the given 3SAT formula be $f$, with variables $X = \{x_1. \ldots , x_n \}$
and clauses $C = \{c_1. \ldots , c_m \}$.
For the reduction, we construct a node set $V$
and transition set \obset{}.

The constructed node set $V$ contains $2n+2$ nodes.
For each variable $x_i \in X$,
$V$ contains the two nodes $y_i$ and $\overline{y_i}$.
Intuitively, node $y_i$ corresponds to the literal $x_i$,
and node $\overline{y_i}$ corresponds to the literal $\overline{x_i}$.
We refer to these $2n$ nodes as {\em literal nodes}.
There are also two additional nodes: $z$ and $z'$.
Transition set \obset{} contains $n+m+2$ transitions, as follows:
$\obset{} = \obset{}^1 \cup  \obset{}^2 \cup \obset{}^3$.

$\obset{}^1$ contains two transitions.  In the first of these
transitions, the predecessor has only node $z$ equal to 1, and the
successor  has all nodes equal to 0.  In the second transition, the
predecessor has only the two nodes $z$ and $z'$ equal to 1, and the
successor has only node $z$ equal to 1.
$\obset{}^2$ contains $n$ transitions, one for each variable in $X$.
For each variable $x_i \in X$,
$\obset{}^2$ contains a transition where
in the predecessor only nodes $y_i$ and $\overline{y_i}$ are equal to 1,
and the successor has all nodes equal to 0.
$\obset{}^3$ contains $m$ transitions, one for each clause of $f$.
For each clause $c_j$ of $f$,
$\obset{}^3$ contains a transition where in the predecessor only node $z$
and the nodes corresponding to the literals that occur in $c_j$
are equal to 1,
and the successor has only node $z$ equal to 1. The construction can be done in polynomial time.
It can be shown that $f$ is satisfiable iff
there exists a threshold-SyDS that is consistent with $\obset$.
\QED

\noindent
\textbf{Remark.} The usefulness of establishing \cnp-hardness results for the $\Gamma$-Consistency problem is pointed out by the next Lemma~\ref{lem:general_hardness}, which states that if $\Gamma$-Consistency is \textbf{NP}-hard, then the hypothesis class for $\Gamma$ SyDSs is \textbf{not} efficiently \pac{} learnable. In the statement of the result, \crp{} denotes the
class of problems that can be solved in randomized polynomial time.  It is widely believed that the complexity classes \cnp{} and \crp{} are different; see \cite{Arora-Barak-2009} for additional details regarding these complexity classes.

\begin{mybox2}
\begin{lemma}\label{lem:general_hardness}
Let $\Gamma$ be any class of SyDS for which $\Gamma$-Consistency is \textbf{NP}-hard. The hypothesis class for $\Gamma$ SyDSs is \textbf{not} efficiently \pac{} learnable, unless \cnp{} = \crp{}.
\end{lemma}
\end{mybox2}

\noindent
\textit{Proof sketch.} If the hypothesis class for $\Gamma$-SyDS admits an efficient \pac{} learner $\Apac{}$, then one can construct an \crp{} algorithm $\Aerm{}$ (based on $\Apac{}$) for the $\Gamma$-Consistency problem, thereby implying \cnp{} = \crp{}. Details of the proof appear in the appendix.
\QED

\noindent
We now present the theorem on the hardness of learning, following Lemmas~\ref{lem:np-hardness} and \ref{lem:general_hardness}.

\begin{mybox2}
\begin{theorem}\label{thm:np_tree_thr_two}
Unless \textbf{NP} = \textbf{RP}, the hypothesis classes for the following classes of SyDSs are \textbf{not} efficiently \pac{} learnable:
$(i)$ \undthr{} and $(ii)$ \treethrtwo{}.
\end{theorem}
\end{mybox2}

%%%% We will have this in the intro section.
%\noindent
%\textit{Remark.} \textcolor{red}{Mention with citations other %learning problems for which hardness results are known.}

\iffalse 
%%%%%%%%%%%%%
\begin{mybox2}
\begin{theorem}
Unless \textbf{NP} = \textbf{RP}, the hypothesis class for \treethrtwo{} is \textbf{not} efficiently \pac{} learnable.
\end{theorem}
\end{mybox2}
%%%%%%%%%%%%%
\fi 

\noindent
\textbf{Sample complexity.}  
For any finite hypothesis class $H$,
given the \pac{} parameters $\epsilon, \delta > 0$, 
the following is a well known 
\cite{haussler1988quantifying}
upper bound on the  sample complexity $\sampcom{H}$ for learning $H$:
$\sampcom{H}\le
\frac{1}{\epsilon}\big(\log(|H|+\log(1/\delta)\big)$.
As mentioned earlier, the size of the hypothesis class $\hclass{}$ is 
$O(2^{\binom{n}{2}} \cdot n^n)$.
Thus,
one can obtain an upper bound on the sample complexity of learning $\hclass{}$: 

\begin{equation}\label{eq:sample_complexity}
\sampcom{\hclass{}} \leq \frac{1}{\epsilon} \cdot \left(n^2 + n \log{}(n) + \log{}(\frac{1}{\delta}) \right)
\end{equation} 
From the above inequality, it follows that $\sampcom{\hclass{}} = O\left((1 / \epsilon) \cdot (n^2 + \log{}(1/\delta))\right)$. In a later section, we will prove a lower bound on the sample complexity which is tight to within a constant factor from the upper bound~(\ref{eq:sample_complexity}), thereby showing that $\sampcom{\hclass{}} = \Theta\left((1 / \epsilon) \cdot (n^2 + \log{}(1/\delta))\right)$.

\smallskip 

\noindent
\textbf{Preview for the results in Sections~\ref{sse:matching} 
and \ref{sse:directed}.}~
In the next subsections, we present
classes of SyDSs which are efficiently
\pac{} learnable. In both cases, we obtain the result by showing that for the corresponding class of SyDSs, the $\Gamma$-Consistency problem is
efficiently solvable.
In other words, given a training set \obset{}, these algorithms represent
\textbf{efficient consistent learners} for the corresponding classes of SyDSs.
As is well known, an efficient consistent learner for a hypothesis class is also
an efficient \pac{} 
learner~\cite{Shwartz-David-2014}.

\subsection{\pac{} Learnability for Matchings}
\label{sse:matching}
% When the graph is a perfect matching, the graph has an even number of nodes and the degree of each node is 1.
Let \matthr{} denote the set of SyDSs where the underlying graph is a perfect matching, and the interaction functions are threshold functions.
In this section, we present an efficient \pac{} learner for the hypothesis class $\hclassmatch{}$, consisting of \matthr{}.  As mentioned earlier, we obtain an efficient \pac{} learner for this class by presenting an efficient algorithm for the $\Gamma$-Consistency problem. We begin with a few definitions.

%\smallskip
\noindent
\textbf{Threshold-Compatibility.} 
For a pair of distinct vertices $u$ and $v$,
we say that $u$ and $v$ are {\em threshold-compatible}
if for all $(C_i, C_i')$ and $(C_j, C_j') \in \bbo$,
if $score(C_i, \{u,v\}) \leq score(C_j, \{u,v\})$,
then $C_i'(u) \leq C_j'(u)$ and $C_i'(v) \leq C_j'(v)$.
%We say that $u$ is {\em threshold-matchable} via $v$ if
%for all $(C_i, C_i') \in \bbo_{0,u}$ and $(C_j, C_j') \in \bbo_{1,u}$,
%$score(C_i, \{u,v\}) < score(C_j, \{u,v\})$.
%We say that $u$ and $v$ are {\em threshold-compatible} if $(i)$ $u$ is threshold-matchable via $v$, and $(ii)$ $v$ is threshold-matchable via $u$.
Informally, $u$ and $v$ are threshold-compatible iff there exist functions $f_u$ for $u$ and $f_v$ for $v$,
each of which is a threshold function of $u$ and $v$, such that $f_u$ and $f_v$ are each consistent with $\bbo{}$.

%\smallskip
\noindent
\textbf{Compatibility Graph.} 
The {\em threshold-compatibility graph} $\G{}' = (\Gv, \Ge')$ of $\obset$ is an undirected graph with vertex set $\Gv{}$, 
and an edge $(u,v) \in \Ge'$ for each pair of threshold-compatible vertices $u$ and $v$. 

\begin{algorithm}[H]
\small
\SetKwInOut{Input}{Input}
\SetKwInOut{Output}{Output}
\SetAlgoLined
\setstretch{0.5}
\Input{The vertex set $\Gv{}$; A training set $\bbo{}$}

\Output{ An \undthr{} SyDS 
$\mathcal{S} = (G, \mc{F})$}
\smallskip

 Construct the threshold-compatibility graph $\G{}' = (\Gv, \Ge')$ of $\bbo{}$ 

 If $\G{}$ does \underline{not} have a perfect matching, answer ``No'' and \textbf{stop}.

 Let $\Ge''$ be the edge set of a perfect matching in $\G{}$.

 $G \gets (\Gv, \Ge'')$; $\mc{F} = \emptyset$ \\
 
 \For{\textbf{each} $v \in \Gv{}$}
 {   
    \eIf{$|\bbo_{1,v}| = 0$}
    {
        $f_v \gets$ the threshold function where $\tau_v = 3$  \\
    }
    {
        let $u$ be the neighbor of $v$ in $G$
        
        $z \gets$ the minimum value of $score(C, \{u,v\})$ over all $(C, C') \in \bbo_{1,v}$

        $f_v \gets$ the threshold function where $\tau_v = z$ 
    }
    $\mc{F} \gets \mc{F} \cup \{f_v\}$
 }
 \Return{$\mc{S} = (G, \mc{F})$}\\
 \caption{\texttt{Full-Infer-Matching}($\Gv{}$)}
 \label{alg:inf-matching}
\end{algorithm}

% Our efficient algorithm for the $\Gamma$-Consistency problem for the class
% of \matthr{} appears as Algorithm~\ref{alg:inf-matching}.
% The algorithm first constructs the threshold-compatibility graph 
% $\G{}' = (\Gv, \Ge')$ of $\obset$.
% The reason for this computation is given in the following lemma.

% \iffalse
% %%%%%%%%%% This definition is not needed.
% \smallskip
% \noindent
% \textbf{Projection.} Given a training set $\obset$ and a subset of vertices $\mc{Y} \subseteq \Gv{}$, we let $\obset[\mc{Y}]$ denote the {\em projection} of $\obset$
% onto vertex set $\mc{Y}$, i.e., a pair $(\mc{C}_\mc{Y}, \mc{C}_\mc{Y}')$ is in $\obset[\mc{Y}]$ iff there is a $(\calc_i,\calc_i') \in \obset$ such that $\mc{C}_\mc{Y} = \calc_i[\mc{Y}]$ and $\mc{C}_\mc{Y}' = \calc_i'[\mc{Y}]$. 
% %%%%%%%%%%%%%
% \fi

%\smallskip
%\noindent
%\textbf{Compatiblility.} A pair of distinct vertices $y$ and $z$ are {\em compatible} if the projection
%$\obset[\{ y, z\}]$ represents a {\em partial function}, i.e., for each $(\calc_i,\calc_i')$ and $(\calc_j,\calc_j')$ in $\obset$,
%$\calc_i[\{ y, z\}] = \calc_j[\{ y, z\}]$ implies that $\calc_i'[\{ y, z\}] = \calc_j'[\{ y, z\}]$. We define the \textbf{compatibility graph} $\G{}' = (\Gv, \Ge{}')$ of $\obset$ to be an undirected graph with the same vertex set $\Gv{}$ as $G$, and an edge $(y,z) \in E'$ for each pair of compatible vertices $y$ and $z$. 

\smallskip
\noindent
\textbf{An efficient learner.} Our efficient algorithm for the $\Gamma$-Consistency problem for the class
of \matthr{} appears as Algorithm~\ref{alg:inf-matching}.
The algorithm first constructs the threshold-compatibility graph 
$\G{}' = (\Gv, \Ge')$ of $\obset$.
The reason for this computation is given in the following lemma (proof in Appendix).

\begin{mybox2}
\begin{lemma}\label{lem:thresh_comp_graph}
The answer to the $\Gamma$-Consistency problem for \matthr{} is ``Yes''
if and only if
the threshold-compatibility graph 
$\G{}' = (\Gv, \Ge')$ of $\obset$ contains a perfect matching.
\end{lemma}
\end{mybox2}

% \noindent
% \textit{Proof.}~ See Appendix.
%Note that each edge in the graph $\G{} = (\Gv{}, \Ge{})$ of the ground-truth system $\mc{S}^*$ is also an edge in $\Ge'$.
%Since $\G$ is a perfect matching, $\G'$ contains a perfect matching. 
%\QED 

\smallskip 

Next, the algorithm finds a maximum matching in $\G{}'$.
Let $\Ge''$ be the edge set of this maximum matching.
Note that, from Lemma~\ref{lem:thresh_comp_graph}, $\G{}'$ contains a perfect matching, so $\Ge''$ is a perfect matching of $\G$.
The learned hypothesis $\cals$ is a \matthr{} on $\Gv{}$ whose graph has edge set $\Ge''$, and interaction function $f_v$ for each vertex $v$, with incident edge $(v,u)$ in $\Ge''$, is any threshold function of variables $v$ and $u$ that is consistent with $\obset[\{ v, w\}]$.

\noindent
\textbf{Remark.} To estimate the running time of 
Algorithm~\ref{alg:inf-matching}, note that for any
pair of nodes, determining compatibility can be done in $O(nq)$ time, where $n$ is the number of nodes and $q$ is
the number of given observations.
Thus, the time to construct the compatibility graph is $O(n^3q)$.
All the other steps (including the computation of perfect matching
which can be done in $O(n^3)$ time~\cite{CLRS-2009}) are dominated by the time to construct the compatibility graph.
Thus, the overall running time is $O(n^3q)$, which is polynomial
in the input size. Hence, we have an efficient consistent learner for the class of \matthr{}.

\iffalse
%%%%%%%%%%%
It can be seen that the algorithm runs in polynomial time. 
Further, the learned system $\cals$ is consistent with the training set $\obset$. 
%(see the Appendix for the full proof). 
It follows that $\cals$ satisfies the $\epsilon$-$\delta$ \pac{} guarantee.
Thus, from Equation~(\ref{eq:sample_complexity}), the hypothesis class for \matthr{} is efficiently \pac{} learnable~\cite{Shwartz-David-2014}. 
%%%%%%%%%
\fi

\begin{mybox2}
\begin{theorem}\label{thm:matching}
The hypothesis class associated with \matthr{} is efficiently \pac{} learnable.
\end{theorem}
\end{mybox2}

\subsection{\pac{} Learnability for Directed Graphs}
\label{sse:directed}

In this section, we present an efficient \pac{} learner for the hypothesis class $\hclassdb{}$ consisting of SyDSs where the underlying graph of the target system is directed, with in-degree bounded by some fixed $\Delta$,
and the interaction functions are threshold functions. As before, we establish this result by presenting an efficient algorithm for the $\Gamma$-Consistency
problem, where $\Gamma$ is the class of SyDSs on directed graphs where the maximum indegree is bounded by a constant $\Delta$ and the interaction functions are threshold functions. We refer to these as  \dthr{}.

\smallskip

\noindent
We say that a given vertex $v$ is {\em threshold-consistent} with a given training set $\bbo{}$ {\em via}
a given vertex set $\mc{Y} \subseteq \Gv{} \, \setminus \{v\}$ if
for all $(C_i, C_i')$ and $(C_j, C_j') \in \bbo$, it holds that if $C_i'(v) < C_j'(v)$, then $score(C_i, \{v\} \, \cup \, \mc{Y}) < score(C_j, \{v\} \, \cup \, \mc{Y})$.
A key lemma that leads to our algorithm is the
following.

% $(i)$ if $score(C_i, \{v\} \, \cup \, \mc{Y}) = score(C_j, \{v\} \, \cup \, \mc{Y})$, then $C_i'(v) = C_j'(v)$, and $(ii)$ if $score(C_i, \{v\} \, \cup \, \mc{Y}) < score(C_j, \{v\} \, \cup \, \mc{Y})$, then $C_i'(v) \leq C_j'(v)$.

\begin{mybox2}
\begin{lemma}\label{lemma:directed}
There exists a $\mc{S} \in \hclassdb{}$ that is consistent with a given training set $\bbo{}$ if and only if every vertex $v$ is threshold-consistent with $\bbo{}$ via a vertex set $N_v$ of cardinality at most $\Delta$.
\end{lemma}
\end{mybox2}

% \noindent
% \textit{Proof.}~ See Appendix.
\iffalse 
%%%%%%%%%
Suppose there exists a system $\mc{S} \in \hclassdb{}$ that is consistent with $\bbo{}$. In this case, $N_v$ is the set of in-neighbors of a vertex $v \in \Gv{}$ in $\mc{S}$. Let $\tau^*_v$ be the threshold of $v$. Consider a partition $\bbo{} = \{\bbo{}_{0,v}, \bbo{}_{1,v}\}$ where $C'(v) = i$, for all $(C, C') \in \bbo{}_{i,v}$, $i = 0, 1$, it follows that $score(\mc{C}, v) < \tau^*_v, \; \forall (\mc{C}, \mc{C}') \in \bbo{}_{0,v}$, and $score(\mc{C}, v) \geq \tau^*_v, \; \forall (\mc{C}, \mc{C}') \in \bbo{}_{1,v}$, where $score(\mc{C}, v)$ is the score of $v$ under $\mc{C}$, with $N$ being the set of in-neighbors of $v$. Thus, we have $\max_{(\mc{C}, \mc{C}') \in \bbo{}_{0,v}} score(\mc{C}, v) < \min_{(\mc{C}, \mc{C}') \in \bbo{}_{1,v}} score(\mc{C}, v)$, and $v$ is threshold consistent.

\par For the other direction, suppose that for each $v \in \Gv{}$, a corresponding subset $N_v \subseteq \Gv{}$ exists. We now argue that there is a system $\mc{S} \in \hclassdb{}$ that is consistent with $\bbo{}$. In particular, $(i)$ the underlying network is a directed graph with $N(v)$ be the set in-neighbors of each vertex $v \in Gv{}$, and $(ii)$ the threshold of each vertex is simply $\min_{(\mc{C}, \mc{C}') \in \bbo{}_{1,v}} score(\mc{C}, v)$. One can easily verify that such a $\mc{S}$ is consistent with $\bbo{}$.
\QED
%%%%%%%%%%
\fi 

\smallskip

\noindent
As shown below, the above lemma provides a straightforward
algorithm for the $\Gamma$-Consistency problem
for the class of \dthr{}.

\begin{mybox2}
\begin{lemma}
For any fixed value $\Delta$, the 
$\Gamma$-Consistency problem for the class of \dthr{}
can be solved efficiently.
\end{lemma}
\end{mybox2}

\noindent
\textit{Proof.} Since the graph is directed, each vertex can be treated independently.
For each vertex $v$, an algorithm can enumerate all possible vertex sets $\mc{Y} \subseteq \Gv{} \, \setminus \{v\}$ of cardinality at most $\Delta$ and find the corresponding $N_v$. The number of such vertex sets is $O(n^{\Delta})$. Further, for each such a set, we check if $v$ is threshold-consistent under this set, which takes $O(\Delta q)$ time. Thus, for each vertex, the time to find an $N_v$ is $O(n^{\Delta} \cdot \delta q)$, and $O(n^{\Delta+1} \cdot \delta q)$ over all vertices.
\QED

\noindent
Since we have an efficient consistent learner
for the hypothesis class $\hclassdb{}$ for any
constant $\Delta$:

\begin{mybox2}
\begin{theorem}
For any fixed value $\Delta$,  the hypothesis class $\hclassdb{}$ is efficiently \pac{} learnable.
\end{theorem}
\end{mybox2}

\section{Partially Observed Networks}\label{sec:partial}
In this section, we consider the learning problem for \undthr{} when the network is {\em partially observed}, with at most $k$ missing edges from the true network~$\G{}$. Let~$\Gobs$ denote the observed network of the system, and let $\hclasspartial{}$ denote the corresponding hypothesis class. The goal is to learn a system in $\hclasspartial{}$ with underlying graph~$G'$ being a supergraph of~$\Gobs$, with at most $k$ additional edges. 

We provide an upper bound on the sample complexity of learning $\hclasspartial{}$ based on a detailed analysis of hypothesis class size. Then, for the setting where at most one edge is missing for each vertex, we present an efficient PAC learner.

\begin{mybox2}
\begin{theorem}\label{thm:missing-k-samp}
Given a partially observed network~$\Gobs$, for any~$\epsilon,\delta > 0$, the sample complexity of learning the hypothesis class $\hclasspartial{}$ satisfies $\sac(\epsilon,\delta) \le
\frac{1}{\epsilon}\big(n\log(\davg(\Gobs)+3)+ck\log (n^2/k) +
\log(1/\delta)\big)$
for some constant~$c>0$, where~$\davg(\Gobs)$ is the average degree of~$\Gobs$.
\end{theorem}
\end{mybox2}

\noindent
\textit{Proof.}
We first bound the size of~$\mc{H}$ of all possible SyDSs in the class \undthr{}, given~$\Gobs$ as the partially observed network. This includes all such SyDSs with the underlying graph being~$\Gobs$ and up to~$k$ more edges. Our sample complexity bound is then based on the following result by
\cite{haussler1988quantifying}:~$\sac(\epsilon,\delta)\le
\frac{1}{\epsilon}\big(\log(|H|+\log(1/\delta)\big)$.

Given a graph~$\G{}$, let~$H_G$ denote the set of threshold SyDS with~$\G{}$ as the underlying graph. From \cite{adiga2018learning}, the size of~$H_G$ can be bounded by accounting for the number of threshold assignments possible for each vertex, and is given by~$|H_G| =
\Pi_{v\in V}\big(d(v)+3\big)\le \big(\davg(G)+3\big)^n$ (See
Theorem~1,~\cite{adiga2018learning}). 

Let~$\gclass(d)$ be the set of
graphs which have the same edge set as~$\Gobs$ plus exactly~$d$ more edges, $d \leq k$;
i.e.,~$G\in\gclass(d)$ iff~$E(G)\supseteq E(\Gobs)$ and~$|E(G)\setminus
E(\Gobs)|=d$.
Let $m$ be the number of edges in $\Gobs$.
For~$G\in\gclass(d)$, note that 
$$\davg(G)=\davg(\Gobs)+d/2n=\davg^*+d/2n$$
where~$\davg^*=\davg(\Gobs)$ for convenience. It follows that the number of such graphs
is
$$|\gclass(d)| = \binom{\binom{n}{2}-m}{d}\le (en^2/d)^{d}$$
using the fact that~$\binom{a}{b}\le \big(\frac{ea}{b}\big)^b$
\cite{Graham-etal-1994}.
Now, the size of the hypothesis class corresponding to threshold SyDS with a
partially observed underlying graph~$\Gobs$ with at most~$k$ edges missing
can be bounded:
\begin{align*}
\sum_{d=0}^k\sum_{G\in\gclass(d)}|H_G| &\le
\sum_{d=0}^k\sum_{G\in\gclass(d)} \big(\davg^*+d/2n+3\big)^n \\
&=\sum_{d=0}^k|\gclass(d)|\big(\davg^*+d/2n+3\big)^n \\
&\le(\davg^*+3)^n\sum_{d=0}^k
\bigg(\frac{en^2}{d}\bigg)^{d}e^{1+\frac{d}{2(\davg^*+3)}} \\
&\le2\big(\davg^*+3\big)^n\sum_{d=0}^k\bigg(\frac{e^2n^2}
{d}\bigg)^{d}\\
&\le c'\big(\davg^*+3\big)^n\bigg(\frac{e^2n^2}{k}\bigg)^{k},
\end{align*}
for some constant~$c'>0$. In particular, the last inequality can be obtained as follows:
$$\sum_{d=0}^k\big(\frac{e^2n^2}{d}\big)^{d} \le
1+\bigintss_1^k\big(\frac{e^2n^2}{x}\big)^xdx$$
Lastly, setting~$y=2x\log(en) -
x\log x$, we have 
\begin{align*}
\bigintss_1^k\big(\frac{e^2n^2}{x}\big)^xdx &=
\bigintss_{x=1}^k\frac{e^y}{2\log(en)-1-\log x}dy \\
&<\bigintss_{x=1}^ke^ydy \le c''e^{k\log(e^2n^2/k)}
\end{align*}

for another constant~$c''>0$. It follows
that~
\begin{align*}
\sac(\epsilon,\delta) &\le
\frac{1}{\epsilon}\big(\log(|H|+\log(1/\delta)\big) \\
&\le
    \frac{1}{\epsilon}\big(n\log(\davg(\Gobs)+3)+ck\log (n^2/k) +
    \log(1/\delta)\big)
\end{align*}
for a suitable constant~$c''>0$.
\QED

\noindent
\textbf{Remark.} For the case where the network is fully known, the work by \cite{adiga2018learning} provides a polynomial-time algorithm that outputs a consistent learner in time~$O(qn)$ where $q$ is the size of $\bbo{}$. In our case, where at most~$k$ edges are missing, the method of considering
all possible supergraphs of~$\Gobs$ with at most~$k$ extra edges and checking the consistency takes~$O\big(n^{2k}pn\big)$ time since there are at most~$n^{2k}$ such graphs. 

By simply setting~$\Gobs$ to be a graph with no edges, Theorem~\ref{thm:missing-k-samp} implies the following corollary when the only information known about the network topology is that it has at most~$m$ edges.

\begin{mybox2}
\begin{corollary}\label{cor:sample_comlexity}
The sample complexity of learning the hypothesis class $\hclass{}$ given that
the underlying network has at most~$m$ edges is $\le
\frac{1}{\epsilon}\big(c\,m\log (n^2/m) + \log(1/\delta)\big)$ for a suitable constant~$c>0$.
\end{corollary}
\end{mybox2}

\subsection{Missing At Most One Edge Per Vertex} 
We now examine the case when the hypothesis class $\hclasspartial{}$ misses at most $k$ edges, of which at most one missing edge is incident on each vertex. We propose an efficient PAC learner for this case. 

\noindent
\textbf{Definitions.} We begin with some definitions.
Consider a given training set $\bbo{}$ and set of vertices $\mc{Y} \subseteq \Gv{}$. 
If $|\bbo{}_{0,v}| = 0$, let $\ell(v, \mc{Y}) = -1$;
otherwise, let $\ell(v, \mc{Y}) = \max_{(\mc{C}, \mc{C}') \in \bbo{}_{0,v}} score(\mc{C}, \mc{Y})$. 
If $|\bbo{}_{1,v}| = 0$, let $h(v, \mc{Y}) = n+1$;
otherwise, let $h(v, \mc{Y}) = \min_{(\mc{C}, \mc{C}') \in \bbo{}_{1,v}} score(\mc{C}, \mc{Y})$. 
Note that the threshold value of $v$ must exceed 
$\ell(v, N^+(\Gobs,v))$.

\noindent
\textbf{Step one.} To infer a SyDS consistent with the observations $\bbo$, we first compute~$\ell(v,N^+(\Gobs,v))$ and~$h(v,N^+(\Gobs,v))$ for each vertex $v$. 
In this process, we also identify all vertices $v$ for which 
$\ell(v,N^+(\Gobs,v)) \geq h(v,N^+(\Gobs,v))$; 
$\Gobs$ violates the threshold consistency condition for each such $v$. 
Let~$\Gv{}'$ denote the set of vertices such that  $\ell(v, N^+(\Gobs,v)) = h(v, N^+(\Gobs,v))$,
and~$\Gv{}''$ denote the set of vertices such that  $\ell(v, N^+(\Gobs,v)) > h(v, N^+(\Gobs,v))$.

\smallskip
\begin{mybox2}
\begin{observation}
Each vertex in $\Gv''$ requires at least two additional incident edges,  so if $\Gv{}'' \neq \emptyset$, 
there is \textbf{no} system in $\hclasspartial{}$ that is consistent with $\bbo$.
\end{observation}
\end{mybox2}

\noindent
So, henceforth we assume that $\Gv{}'' = \emptyset$.
%and |\Gv{}'| \leq 2k$ 

\smallskip
\noindent
\textbf{Step two.} Next, we construct a maximum-weighted matching problem instance
with vertex set $\Gv{}$, where the edge weights are all positive integers.
In particular, we say that a vertex pair $(u,v)$ is {\em viable} if $(i)$ $u \neq v$, 
$(ii)$ $u \in \Gv{}'$ or $v \in \Gv{}'$, and $(iii)$ adding the edge $(u,v)$ to $\Gobs$ would result in $u$ and $v$ both satisfying the consistency condition, i.e.,
$\ell\big(u,N^+(\Gobs,u)\cup\{v\}\big) < h\big(u,N^+(\Gobs,u)\cup\{v\}\big)$
and $\ell\big(v,N^+(\Gobs,v)\cup\{u\}\big) < h\big(v,N^+(\Gobs,v)\cup\{u\}\big)$.
\smallskip

Let $t = |\Gv{}'|$. The constructed graph $\Gm{}$ has an edge for each viable vertex pair. Let $\Ge{}_1$ denote the edges in $\Gm{}$ with exactly one endpoint in $\Gv{}'$, and $\Ge{}_2$ the edges in $\Gm{}$ with both endpoints in $\Gv{}'$.
The edges in $\Ge{}_1$ are given weight $t$, and the edges in $\Ge{}_2$ are given weight $2t+1$.

\smallskip
\noindent
\textbf{Step three.} Lastly, the constructed matching problem is solved, producing a maximum weight matching, $\M$. If $\M$ matches all the vertices in~$\Gv{}'$ and consists of at most $k$ edges, then we construct the new graph~$\G'$ by adding the edges in $\M$
to $\Gobs$. 
Since all added edges are viable, and each vertex is the endpoint of at most one added edge, we have that for all~$v \in \Gv{}$, $\ell\big(v, N^+(\G{}',v)\big) < h\big(v,N^+(\G{}',v)\big)$. 
For each vertex $v$, we set the threshold~$\tau'_v$ to be an integer such that $\ell\big(v,N^+(\G{}',v)\big) < \tau'_v \leq h\big(v,N^+(\G{}',v)\big)$. 
%We return this threshold SyDS as output. 
If the maximum weight matching does not match all vertices in~$\Gv{}'$ or contains more than $k$ edges, then there is no SyDS in $\hclasspartial{}$ that is consistent with the training set $\bbo{}$.

%\noindent
%\textit{Remark.} To see the correctness of the algorithm, one can verify that any matching that includes all vertices in $\Gv{}'$ has weight at least $kt$. On the other hand, any matching that misses at least one vertex in $\Gv{}'$ has weight at most $q^2 - q$. Therefore, if there is a matching that includes all vertices in $\Gv{}'$, then any maximum matching found by the algorithm matches everyone in $\Gv{}'$.

\smallskip
\noindent
\textbf{Correctness.} 
Consider a matching $\M'$ within $G_m$.
Let $\mu(\M')$ denote the number of vertices in $V'$ that are covered by $\M'$, and $W(\M')$ denote the weight of $\M'$. Suppose that $\mu(\M')$ contains $e_1$ edges from $E_1$ and $e_2$ edges from $E_2$.
Then $\mu(\M') = e_1+2e_2$ and $W(\M') = qe_1+2qe_2+e_2$.
Since $e_2 \leq q/2 < q$, $\mu(\M') = \lfloor W(\M')/q \rfloor$.
Thus, no other matching matches more vertices in $V'$ than $\M$.
Moreover, of those matchings that match the same number of $V'$ vertices as $\M$, none has more edges from $E_2$, so none consists of fewer edges than $\M$.
Thus, $\M$ matches as many vertices from $V'$ as possible, and does so with the minimum number of edges possible.

Note that both the $(i)$ construction of the matching graph $\mc{G}_m$ and $(ii)$ finding a maximum matching in $\mc{G}_m$ can be done in polynomial time. It follows that the learning problem considered is efficiently \pac{} learnable.

\begin{mybox2}
\begin{theorem}\label{thm:consistent_learner}
Suppose~$\Gobs(V,E)$ is missing at most $k$ edges, with at most one is incident on each vertex. The corresponding hypothesis class is efficiently PAC learnable.
%in time~$O\big(pm+pnk + n^{3/2}k)\big)$, where~$n$ and~$m$ are the number of vertices and edges in~$\Gobs$.
\end{theorem}
\end{mybox2}

\section{Tight Bounds on the Natarajan Dimension}\label{sec:nata}
In this section, we study the {\em expressiveness} of the hypothesis class $\hclass{}$ for \undthr{}, measured by the Natarajan dimension~\cite{natarajan1989learning} $\Ndim{\hclass{}}$. Specifically, a higher value of $\Ndim{\hclass{}}$ implies a greater expressive power of the class $\hclass{}$. Further, $\Ndim{\hclass{}}$ characterizes the sample complexity of learning $\hclass{}$. 

% We establish a tight bound on $\Ndim{\hclass{}}$. In particular, based on a construction method, we show that the Natarajan dimension of the hypothesis class $\hclass{}$ is at least $n^2/4$. Further, our lower bound is only a constant factor smaller than the upper bound on $\Ndim{\hclass{}}$. Recall that in the definition of shattering, each $\mc{C}$ in a shatterable set $\mc{R}$ is associated with two configurations, denoted by $\mc{C}^A$ and $\mc{C}^B$, where $\mc{C}^A \neq \mc{C}^B$. We first present a key definition:

% \begin{mybox2}
% \begin{definition}[\textbf{Contested Vertices}]\label{def:contested}
% We call a vertex $v$ \textbf{contested} for a $\mc{C} \in \mc{R}$ if $\mc{C}^A(v) \neq \mc{C}^B(v)$.
% \end{definition}
% \end{mybox2}
% \midvs{}

% Each configuration in a shatterable set has at least one contested vertices. In proving a set to be shatterable, it is critical to choose the right vertices to be contested. Theorem~\ref{thm:nata} presents our main result.

\begin{mybox2}
\begin{theorem}\label{thm:nata}
The Natarajan dimension of the hypothesis class $\hclass{}$ is $\geq \lfloor n^2/4 \rfloor$, irrespective of the graph structures. 
\end{theorem}
\end{mybox2}

\noindent
\textit{Proof.} 
We establish the result by specifying a shattered set $\mc{R} \subset \mc{X}$ of size $\lfloor n^2/4 \rfloor$. 
Let the set $\Gv{}$ of $n$ vertices be partitioned into two subsets: $\mc{Y}$ consisting of $\lfloor n/2 \rfloor$ vertices, and $\mc{Z}$ consisting of the other $\lceil n/2 \rceil$ vertices.
Set $\mc{R}$ consists of $|\mc{Y}| \cdot |\mc{Z}| = \lfloor n^2/4 \rfloor$ configurations, as follows.
Each configuration in $\mc{R}$ has exactly two vertices in state 1, one of which is in $\mc{Y}$, and the other in $\mc{Z}$.

Let $g_1$ be the function that maps each configuration $\mc{C}$
into the configuration where each vertex in $\mc{Y}$ has the same state as in $\mc{C}$, and each vertex in $\mc{Z}$ has state 0.
Let $g_2$ be the function that maps each configuration into the configuration where every vertex has state 0.

We now show that the two requirements for shattering are satisfied.
For requirement $(i)$, for each $\mc{C} \in \mc{R}$,  the state-1 vertex in $\mc{Y}$ under $\mc{C}$ is in state 1 in $g_1(\mc{C})$ and in state 0 in $g_2(\mc{C})$, so $g_1(\mc{C}) \neq g_2(\mc{C})$.
For requirement $(ii)$, consider each subset $\mc{R}' \subseteq \mc{R}$.
Let $\mc{S}_{\mc{R}'} = (\mc{G}_{\mc{R}'}, {\cal F}_{\mc{R}'}) \in \mc{H}$ be the following SyDS.
Graph $\mc{G}_{\mc{R}'}$ is a bipartite, between $\mc{Y}$ and $\mc{Z}$, containing an edge $(y,z)$ iff there is configuration in $\mc{R}'$ in which $y$ and $z$ are both in state 1.
Each interaction function is a threshold function,
where the threshold of every vertex in $\mc{Y}$ is 2, and the threshold of every vertex in $\mc{Z}$ is 3.
We now claim that $\forall \mc{C} \in \mc{R}', \; \mc{S}(\mc{C}) = g_1(\mc{C})$ and $\forall \mc{C} \in \mc{R} \setminus \mc{R}', \; \mc{S}(\mc{C}) = g_2(\mc{C})$. 
Consider any $\mc{C} \in \mc{R}$.
Let $y \in \mc{Y}$ and $z \in \mc{Z}$ be the two vertices that are in state 1 in $\mc{C}$.
If $\mc{C} \in \mc{R}'$, then $\mc{G}_{\mc{R}'}$ contains the edge $(y,z)$,
so $\mc{S}(\mc{C}) = g_1(\mc{C})$.
If $\mc{C} \in \mc{R} \setminus \mc{R}'$, then $y$ and $z$ are not neighbors in $\mc{G}_{\mc{R}'}$,
so $\mc{S}(\mc{C}) = g_2(\mc{C})$.
This completes the proof of the claim.
\QED

\noindent
By~\cite{Shwartz-David-2014}, the sample complexity of learning $\hclass{}$ is at most:
\begin{equation}\label{eq:shai}
c_1 \cdot (1/\epsilon) \cdot \left( \Ndim{\hclass{}} + \log{}(1/\delta) \right)
\end{equation}
\noindent
The corollary follows:

\begin{mybox2}
\begin{corollary}\label{coro:sample}
The sample complexity of learning $\mathcal{H}$ satisfies: $\sampcom{\hclass{}} \geq c_1 \cdot (1/\epsilon) \cdot \left( n^2/4 + \log{}(1/\delta) \right)$
\end{corollary}
\end{mybox2}

\noindent
\textbf{Remark.} For fixed $\epsilon$, $\delta$, 
Equation~\eqref{eq:shai} shows that the sample complexity is $\Omega(\Ndim{\hclass{}})$. Further, Equation~(\ref{eq:sample_complexity}) states that the sample complexity of learning $\hclass{}$ is $O(n^2)$. It follows that $\Ndim{\hclass{}} = O(n^2)$, and our lower bound of $\lfloor n^2/4 \rfloor$ is only a constant factor away from the lowest upper bound on $\Ndim{\hclass{}}$.

\begin{mybox2}
\begin{corollary}
The Natarajan dimension of the hypothesis class $\hclass{}$ is $\leq c \cdot n^2$ for some constant $c$.
\end{corollary}
\end{mybox2}

\section{Conclusion and Future Work}
\label{sec:concl}
We examined the problem of learning both the topology and the interaction functions of an unknown networked system.
We showed that the problem in general is computationally intractable. We then identified special classes that are efficiently solvable. Further, we studied a setting where the underlying network is partially observed, 
and proposed an efficient \pac{} algorithm. 
%There are several directions for future work. 
%For example, 
It would be interesting to extend
our efficient algorithms to the case where the
observation set includes both positive and negative
examples of transitions. It would also be of interest to consider the problem where additional information about the graph (e.g., maximum node degree, the size of a maximum clique) and/or the interaction functions (e.g., upper bounds on the threshold values) is also available. 

% ----------------
%       Bib      -
% ----------------
\bibliographystyle{plain}
\bibliography{bib}

\clearpage

% Appendix
\begin{center}
\fbox{{\Large\textbf{Appendix}}}
\end{center}

\section{Additional Material for Section~\ref{sec:general_complexity}}
\label{sec:app_general_complexity}

\subsection{Notation Used in the Paper}
\label{app_see:notation}

\begin{table}[h]
\begin{center}
\begin{tabular}{|c|p{4.5in}|}\hline
\textbf{Symbol} & \textbf{Explanation} \\ \hline\hline
$\G(\Gv, \Ge)$ &
  {Underlying graph of a SyDS with vertex set $\Gv$ and edge set $\Ge$} \\ \hline
$n$ & {Number of nodes in the underlying graph}  \\ \hline
$d(v)$ & {Degree of node $v$} \\ \hline
$\davg(G)$ & {Average node degree of graph $G$} \\ \hline
$\Delta$ & {Maximum indegree of a directed graph}  \\ \hline
$\mathcal{F}$ & {Set of local interaction functions} \\ \hline
$\calc$ & {A SyDS configuration} \\ \hline
$\chi$ & {The set of all $2^n$ (Boolean) configurations over $n$ nodes} \\ \hline
$\mathcal{S}^*$ & {Ground-Truth SyDS} \\ \hline
$\obset{}$ & {Observation (i.e., training) set with configuration
              pairs  of the form $(\calc,\calcp)$,
              where \calcp{} is the successor of \calc{}}  \\ \hline
$q$ & {No. of observations (i.e., $|\obset{}|$)} \\ \hline
$\delta$, $\epsilon$ & {Parameters associated with the \pac{} model} \\ \hline
$\Gobs$ & {Partially observed graph} \\ \hline
$\hclass$ & {Hypothesis class associated with \undthr{}} \\ \hline
$\hclassdb$ & {Hypothesis class associated with \dthr{}} \\ \hline
$\hclasspartial$ & {Hypothesis class associated with partially observed graph} \\ \hline
$m_H$ & {Sample complexity of hypothesis class $H$} \\ \hline
\Ndim{$H$} & {Natarajan dimension of hypothesis class $H$} \\ \hline
\end{tabular}
\caption{Symbols used in the paper and their interpretation}
\label{tab:symbol_def}
\end{center}
\end{table}

\subsection{Classes of SyDSs Considered}
\label{app_sse:syds_classes}

\begin{table}[tb!h]
%{\small
\begin{center}
    \begin{tabular}{|l|p{3.95in}|}\hline
    \textbf{SyDS Notation} & \textbf{Description}\\ \hline\hline
     
     \undthr{} & {The graph is undirected and the interaction functions are threshold functions.} \\ \hline

     \treethrtwo{}{} & {The graph is an undirected tree and the interaction functions are 2-threshold functions.} \\ \hline

      \matthr{} & {The graph is a perfect matching and the interaction functions are threshold functions.} \\ \hline  

      \dthr{} & {The graph is directed, with in-degree bounded by some fixed $\Delta$, and the interaction functions are threshold functions.} \\ \hline  
    \end{tabular}
\end{center}
%}
\caption{Notation for Classes of SyDS}
\label{tab:syds_class}
\end{table}

\clearpage  

\subsection{Statement and Proof of Lemma~\ref{lem:np-hardness}}

\medskip

\begin{mybox2}
\textbf{Lemma~\ref{lem:np-hardness}.} {\em The $\Gamma$-Consistency problem is \textbf{NP}-complete for
the following classes of SyDSs: $(i)$ \undthr{} and $(ii)$ \treethrtwo{}.}
\end{mybox2}

\noindent
\textit{(i) Proof for \undthr{}.} Given a SyDS
from the class of \undthr{}, one can easily verify if it is consistent with a transition set \obset{} in polynomial time; thus, the problem is in $\cnp$. The proof of $\cnp$-hardness is via a reduction from 3SAT. Let the given 3SAT formula be $f$, with variables $X = \{x_1. \ldots , x_n \}$ and clauses $C = \{c_1. \ldots , c_m \}$. For the reduction, we construct a vertex set $\Gv{}$  and training set \obset{} as follows.

The constructed vertex set $\Gv{}$ contains $2n+2$ vertices.
For each variable $x_i \in X$,
$\Gv{}$ contains the two vertices $y_i$ and $\overline{y_i}$.
Intuitively, vertex $y_i$ corresponds to the literal $x_i$,
and vertex $\overline{y_i}$ corresponds to the literal $\overline{x_i}$.
We refer to these $2n$ vertices as {\em literal vertices}.
There are also two additional vertices: $z$ and $z'$. Transition set \obset{} contains $n+m+2$ transitions, defined as $\obset{} = \obset{}^1 \cup  \obset{}^2 \cup \obset{}^3$ where:

\begin{description}[noitemsep,topsep=0pt]
\item{$(1)$} $\obset{}^1$ contains two transitions.  In the first of these transitions, the predecessor has only vertex $z$ equal to 1, and the successor has all vertices equal to 0.  In the second transition, the
predecessor has only the two vertices $z$ and $z'$ equal to 1, and the
successor has only vertex $z$ equal to 1.
\item{$(2)$} $\obset{}^2$ contains $n$ transitions, one for each variable in $X$. For each variable $x_i \in X$,
$\obset{}^2$ contains a transition where in the predecessor only vertices $y_i$ and $\overline{y_i}$ are equal to 1,
and the successor has all vertices equal to 0.

\item{$(3)$} $\obset{}^3$ contains $m$ transitions, one for each clause of $f$. For each clause $c_j$ of $f$,
$\obset{}^3$ contains a transition where in the predecessor only vertex $z$ and the vertices corresponding to the literals that occur in $c_j$
are equal to 1,
and the successor has only vertex $z$ equal to 1.
\end{description}

We now show that $f$ is satisfiable if and only if there exists a threshold-SyDS that is consistent with $\obset$. First, assume that $f$ is satisfiable. Let $\alpha$ be a satisfying assignment for $f$.
Let $\cals(\alpha)$, 
denote the following threshold-SyDS over $\Gv{}$.
The edge set of $\cals(\alpha)$ is as follows.
For each variable $x_i$ such that $\alpha(x_i) = 1$,
there is an edge between vertex $z$ and vertex $y_i$.
For each variable $x_i$ such that $\alpha(x_i) = 0$,
there is an edge between vertex $z$ and vertex $\overline{y_i}$.
There is also an edge between vertices $z$ and $z'$.
There are no other edges.
The vertex function $f_z$ of vertex $z$ is the threshold function with threshold 2, 
and every other vertex function is the constant function 0. By examining the transitions in \obset{},
it can be verified that $\cals(\alpha)$ is consistent with \obset{}.
Crucially, for each given transition in $\obset{}^3$, 
corresponding to a given clause of $f$, 
the predecessor configuration has at least two generalized neighbors of $z$ 
equal to 1: vertex $z$ and a literal vertex corresponding to a literal that
is made true by $\alpha$. 

Now assume that there exists a threshold-SyDS that is consistent with $\obset$.
Let $\cals$ be such a threshold-SyDS.
Let $\alpha(\cals)$ denote the assignment to variables $X$ such
that for each $x_i \in X$, $x_i$ has value 1 iff the underlying
graph of $\cals$ contains an edge between vertex $z$ and vertex $y_i$. We claim that $\alpha(\cals)$ is a satisfying assignment for $f$: $(i)$ From the two transitions in $\obset^1$, $\tau_z = 2$. $(ii)$ From the $n$ transitions in $\obset^2$, for each $x_i \in X$,
since $\tau_z = 2$, there is an edge between vertex $z$ and at most one of $y_i$ and $\overline{y_i}$. $(iii)$ From the $m$ transitions in $\obset^3$, for each clause $c_j \in C$, since $\tau_z = 2$, there is an edge between $z$ and at least one of the literal vertices corresponding to the literals that occur in $c_j$.
Thus, $\alpha(\cals)$ is a satisfying assignment for $f$. This concludes the proof. \QED

\smallskip 

\noindent
\textit{(ii) Proof for \treethrtwo{}.} Here also, it is
easy to verify that the problem is in \cnp. 
The proof of $\cnp$-hardness is via a reduction from 3SAT. 
Let the given 3SAT formula be $f$, with variables $X = \{x_1. \ldots , x_n \}$
and clauses $C = \{c_1. \ldots , c_m \}$.
For the reduction, we construct a node set $V$ and transition set \obset{}, as follows.

The constructed node set $V$ contains $4n+3$ nodes.
For each variable $x_i \in X$,
$V$ contains the four nodes $y_i$, $\overline{y_i}$, $w_i$ and $w'_i$.
Intuitively, node $y_i$ corresponds to the literal $x_i$,
and node $\overline{y_i}$ corresponds to the literal $\overline{x_i}$.
We refer to the $2n$ nodes of the form $y_i$ and $\overline{y_i}$ 
as \textbf{literal nodes}.
There are also three additional nodes: $z$, $z'$ and $z''$.

Transition set \obset{} contains $4n+m+3$ transitions, as follows. For simplicity, when we specify configurations, we indicate only those nodes whose states are 1.

\begin{description}[noitemsep,topsep=0pt]

\item{$(1)$} $\obset{} = \obset{}^1 \cup \obset{}^2 \cup \obset{}^3 
\cup \obset{}^4 \cup \obset{}^5$.

\item{$(2)$} $\obset{}^1$ consists of the following transition,
which is the only transition that is not a fixed point transition: $z \rightarrow \emptyset$.

\item{$(3)$} $\obset{}^2$ consists of the following two transitions:
$z \; z' \; \rightarrow \; z \; z'$ and $z' \; z'' \; \rightarrow \; z' \; z''$.

\item{$(4)$} $\obset{}^3$ contains $2n$ transitions, as follows. 
For each variable $x_i \in X$,
$\obset{}^3$ contains the two transitions
$w_i \; w_i' \; \rightarrow \; w_i \; w_i'$ and 
$w_i \; w_i' \; z' \; z'' \; \rightarrow \; w_i \; w_i' \; z' \; z''$.

\item{$(5)$} $\obset{}^4$ contains $2n$ transitions, as follows. 
For each variable $x_i \in X$,
$\obset{}^4$ contains the two transitions
$w_i \; y_i \; \overline{y_i} \; \rightarrow \; w_i \; y_i \; \overline{y_i}$ and 
$w_i \; y_i \; \overline{y_i} \; z \; \rightarrow 
\; w_i \; y_i \; \overline{y_i} \; z$.

\item{$(6)$} $\obset{}^5$ contains $m$ transitions, one for each clause of $f$. 
For each clause $c_j$ of $f$,
there is a fixed point transition where the only nodes that equal 1 are node $z$,
the nodes corresponding to the literals that occur in $c_j$,
and each node $w_i$ such that the literal $x_i$ or $\overline{x_i}$ occurs in $c_j$.
\end{description}

We will show that the reduction has the following two properties,
both of which are based on 
a relationship between variables that equal 1 in an assignment to $X$,
and SyDS edges from nodes for positive literals to node $z$, as described below.

\noindent
\textbf{Property 1:} if $\alpha$ is a satisfying assignment for $f$,
then there is a SyDS in the class \treethrtwo{} 
such that $\cals$ is consistent with transition set \obset{}.

\noindent
\textbf{Property 2:} 
if  there exists a SyDS $\cals\in$ \treethrtwo{} that is consistent with \obset{}, 
then there is a satisfying assignment for $f$.

\iffalse
%%%%%%%%%%%%%%%%
Since $\Gamma_{UT,\tau = 2} \subseteq \Gamma$, 
Property 1 ensures that if $f$ is satisfiable,
then the answer to the $\Gamma$-Consistent SyDS Problem for \obset{} is $\yes$.
Since $\Gamma \subseteq \Gamma_{D,S}$,
Property 2 ensures that if 
the answer to the $\Gamma$-Consistent SyDS Problem for \obset{} is $\yes$,
then $f$ is satisfiable.
Thus, the two properties together ensure that $f$ is satisfiable
iff the answer to the $\Gamma$-Consistent SyDS Problem for \obset{} is $\yes$,
thereby establishing the $\cnp$-hardness of the $\Gamma$-Consistent SyDS Problem.
%%%%%%%%%%%%%%%%
\fi

\smallskip
\noindent
\textbf{Proof of Property 1}: 
For any assignment $\alpha$ to $X$, let $\cals(\alpha) \in$ \treethrtwo{}
denote the following SyDS over $V$.
Every local function is a threshold function with threshold 2.
The edge set of $\cals(\alpha)$ is as follows.
There are the two edges $(z,z')$ and $(z,z')$.
For each variable $x_i$,
there are the three edges $(w_i, w_i')$, $(w_i, y_i)$ and $(w_i, \overline{y_i}$),
plus an additional edge as follows.
If $\alpha(x_i) = 1$,
then the additional edge for $x_i$ is $(y_i,z)$;
and if $\alpha(x_i) = 0$,
then the additional edge for $x_i$ is $(\overline{y_i},z)$.
There are no other edges.
It can be seen that the underlying graph of $\cals(\alpha)$ is a tree. By examining the transitions in \obset{}, 
it can be verified that $\cals(\alpha)$ is consistent with \obset{}.

\noindent
{\textsf{Claim 1:}}~ If $\alpha$ is a satisfying assignment for $f$,
then SyDS $\cals(\alpha)$ is consistent with \obset{}. 

\smallskip
\noindent
\textbf{Proof of Property 2}: For any SyDS $\cals$ over $V$, where
$\cals \in$ \treethrtwo{}, we let $\alpha(\cals)$ denote the assignment
to variables $X$ such that for each $x_i \in X$, $x_i$ has value 1
iff the underlying graph of $\cals$ contains an edge from node $y_i$
to node $z$.

\noindent
{\textsf{Claim 2:}}~ 
If  SyDS $\cals \in$ \treethrtwo{} with node set $V$ is consistent with \obset{}, 
then $\alpha(\cals)$ is a satisfying assignment for $f$.

For each node $v \in V$, let $T_v$ denote the symmetric table for $f_v$. From the transition $z \rightarrow \emptyset$ in $\obset^1$,  $T_z[1] = 0$. Then, from the transition $z \; z' \; \rightarrow \; z \; z'$ in $\obset^2$, 
$T_z[2] = 1$, and $\cals$ contains an edge from $z'$ to $z$.
Moreover, from the transition $z' \; z'' \; \rightarrow \; z' \; z''$ in $\obset^2$, $\cals$ does not contain an edge from $z''$ to $z$. Next, consider the four nodes for each variable $x_i \in V$.
From the transition $w_i \; w_i' \; \rightarrow \; w_i \; w_i'$ in $\obset^3$,  
at most one of $w_i$ and $w_i'$ is the source of an incoming edge to $z$.
Consequently, from the transition 
$w_i \; w_i' \; z' \; z'' \; \rightarrow \; w_i \; w_i' \; z' \; z''$ in $\obset^3$,  
neither $w_i$ nor $w_i'$ is the source of an incoming edge to $z$.

Next, from the transition 
$w_i \; y_i \; \overline{y_i} \; \rightarrow \; w_i \; y_i \; \overline{y_i}$ 
in $\obset^4$,
at most one of $y_i$ and $\overline{y_i}$ is the source of an incoming edge to $z$.
Moreover, from the transition 
$w_i \; y_i \; \overline{y_i} \; z \; \rightarrow 
\; w_i \; y_i \; \overline{y_i} \; z$ in $\obset^4$, 
at least one of $y_i$ and $\overline{y_i}$ is the source of an incoming edge to $z$.
Thus, exactly one of $y_i$ and $\overline{y_i}$ is the source of an incoming edge to $z$. Finally, for each clause $c_j$ of $f$,
from the $\obset^5$ transition corresponding to that clause,
since $T_z[1] = 0$,
there is an incoming edge to $z$ from at least one of the literal nodes 
corresponding to the literals that occur in $c_j$.  
Thus, $\alpha(\cals)$ is a satisfying assignment for $f$.
\QED

\subsection{Statement and Proof of 
Lemma~\ref{lem:general_hardness}}
\label{app_sse:np_hardness}

\medskip

\begin{mybox2}
\textbf{Lemma~\ref{lem:general_hardness}.}
{\em Let $\Gamma$ be any class of SyDS for which $\Gamma$-Consistency is \textbf{NP}-hard. The hypothesis class for $\Gamma$ SyDSs is \textbf{not} efficiently \pac{} learnable, unless \cnp{} = \crp{}.}
\end{mybox2}

\noindent
\textit{Proof.}~
Let $\Gamma$ be a class of SyDS where $\Gamma$-Consistency is \textbf{NP}-hard. Suppose the hypothesis class
associated with $\Gamma$ SyDS is efficiently \pac{} learnable. Let $\Apac{}$ be an efficient \pac{} learning algorithm whose running time is polynomial in the size of the problem instance, $\delta$, and $\epsilon$.  We
show that $\Apac{}$ can be used to devise an \crp{} algorithm for $\Gamma$-Consistency. Given an instance $\Ierm{}$ of $\Gamma$-Consistency consisting of the training set \bbo{} of transitions, we construct an instance $\Ipac{}$ of the \pac{} learning problem as follows:

\begin{description}[noitemsep,topsep=0pt]
\item{$(1)$} The training set is
the same as the set \bbo.

\item{$(2)$} The distribution $\mathcal{D}$ is defined such that each configuration in \bbo{} is chosen with probability $1 / |\bbo{}|$, and all other configurations are chosen with probability $0$.

\item{$(3)$} Let $\epsilon = 1/(2|\bbo{}|)$ ~and~ $\delta$ = $0.1$.
\end{description}

\noindent
Using the assumed efficient learning algorithm $\Apac{}$, we now present an \crp{} algorithm $\Aerm{}$ for $\Ierm{}$:

\begin{description}[noitemsep,topsep=0pt]
\item{$(1)$} Run the algorithm $\Apac{}$ on the instance $\Ipac{}$.  

\item{$(2)$} If $\Apac{}$ produces a hypothesis (i.e., a $\Gamma$-SyDS)
that is consistent will all the transitions in \bbo{}, 
then $\Aerm{}$ outputs ``Yes"; otherwise, $\Aerm{}$ outputs ``No".
\end{description}

\noindent
Since \Apac{} runs in polynomial time, \Aerm{} also runs in polynomial
time. We first show that a hypothesis $h$ is consistent with \bbo{} if and only if $h$ has an error (over the distribution $\mathcal{D}$) at most $\epsilon$.

\noindent
{\textsf{Claim 1:}}
A hypothesis $h$ is consistent with all the examples in \bbo{} if and only if $h$ incurs an error (over the distribution $\mathcal{D}$) at most $\epsilon = 1/(2|\bbo{}|)$.

\noindent
If a hypothesis $h$ is consistent with \bbo{}, then clearly its error (i.e., the probability of $h$ making a wrong prediction for a configuration drawn from the distribution $\mathcal{D}$) is $0$. Now suppose $h$ is not consistent with \bbo{}; that is, $h$ errs on one or more of the transitions in \bbo{}.
Since the distribution $\mathcal{D}$ is uniform over \bbo{}, the error of $h$ is at least $1/|\bbo|$ which \emph{exceeds} the allowed error $\epsilon = 1/(2|\bbo{}|)$. The claim follows.

\noindent
We now present the next claim, which states that \Aerm{} is an \crp{} algorithm for the problem $\Gamma$-Consistency.

\noindent
{\textsf{Claim 2:}} If the instance $\Ierm{}$ has a solution, then Algorithm
\Aerm{} returns ``Yes" with probability at least $0.9$; otherwise, Algorithm \Aerm{} returns ``No".

\noindent
We proceed with the proofs for the two parts of the claim. Suppose there is a solution to the instance $\Ierm{}$, that is, there is a hypothesis $h$ that is consistent with all the examples in \bbo. By Claim~A.1., such a hypothesis $h$ has an error (over the distribution $\mathcal{D}$) at most $\epsilon$, and the \pac{} algorithm \Apac{} produces such an $h$ with probability at least $1 - \delta = 0.9$. On the other hand, if there is no solution to the instance $\Ierm{}$, then clearly $\Apac{}$ never learns an appropriate hypothesis and $\Aerm{}$ always returns ``No''. This concludes the proof. The theorem follows. \QED

\noindent
\subsection{Statement and Proof of Lemma~\ref{lem:thresh_comp_graph}}
\label{app_sse:thresh_comp_graph_def}

\medskip

\begin{mybox2}
\textbf{Lemma~\ref{lem:thresh_comp_graph}:}
The answer to the $\Gamma$-Consistency problem for \matthr{} is ``Yes''
if and only if
the threshold-compatibility graph 
$\G{}' = (\Gv, \Ge')$ of $\obset$ contains a perfect matching.
\end{mybox2}

\noindent
\textit{Proof.} 
For the \textbf{if} part of the claim,
suppose that edge set $E'$
is a perfect matching within the threshold compatibility graph.
Let $\cals$ be the SyDS on $V$ whose underlying graph has edge set $E'$,
and whose local function $f_y$ for each node $y$,
with with incident edge $(y,z)$ in $E'$,
is any threshold function of variables $y$ and $z$ that is consistent with
the partial function occurring in $\obset[\{ y, z\}]$.
It can be seen that $\cals$ is consistent with $\obset$.
For the \textbf{only if} part of the claim,
suppose that $\cals$ is consistent with $\obset$
and the underlying graph $G=(V,E)$ of $\cals$ is a perfect matching.
Then, $E$ is a perfect matching within the compatibility graph. \QED

\noindent
\subsection{Statement and Proof of 
Lemma~\ref{lemma:directed}}

\medskip

\begin{mybox2}
\textbf{Lemma~\ref{lemma:directed}:}~
There exists a $\mc{S} \in \hclassdb{}$ that is consistent with a given training set $\bbo{}$ if and only if  every vertex $v$ is threshold-consistent with $\bbo{}$ via a vertex set $N_v$ of cardinality at most $\Delta$.
\end{mybox2}

\noindent
\textit{Proof.}~
Suppose there exists a system $\mc{S} \in \hclassdb{}$ that is consistent with $\bbo{}$. In this case, $N_v$ is the set of in-neighbors of a vertex $v \in \Gv{}$ in $\mc{S}$. Let $\tau^*_v$ be the threshold of $v$. Consider a partition $\bbo{} = \{\bbo{}_{0,v}, \bbo{}_{1,v}\}$ where $C'(v) = i$, for all $(C, C') \in \bbo{}_{i,v}$, $i = 0, 1$, it follows that $score(\mc{C}, v) < \tau^*_v, \; \forall (\mc{C}, \mc{C}') \in \bbo{}_{0,v}$, and $score(\mc{C}, v) \geq \tau^*_v, \; \forall (\mc{C}, \mc{C}') \in \bbo{}_{1,v}$, where $score(\mc{C}, v)$ is the score of $v$ under $\mc{C}$, with $N$ being the set of in-neighbors of $v$. Thus, we have $\max_{(\mc{C}, \mc{C}') \in \bbo{}_{0,v}}\{score(\mc{C}, v)\} < \min_{(\mc{C}, \mc{C}') \in \bbo{}_{1,v}} \{score(\mc{C}, v)\}$, and $v$ is threshold consistent.

\par For the other direction, suppose that for each $v \in \Gv{}$, a corresponding subset $N_v \subseteq \Gv{}$ exists. We now argue that there is a system $\mc{S} \in \hclassdb{}$ that is consistent with $\bbo{}$. In particular, $(i)$ the underlying network is a directed graph with $N(v)$ be the set in-neighbors of each vertex $v \in Gv{}$, and $(ii)$ the threshold of each vertex is simply $\min_{(\mc{C}, \mc{C}') \in \bbo{}_{1,v}} \{score(\mc{C}, v)\}$. One can easily verify that such a $\mc{S}$ is consistent with $\bbo{}$.
\QED

\end{document}